\documentclass[sigconf,nonacm]{acmart}
\usepackage{graphicx}
\usepackage{makecell}
\usepackage[normalem]{ulem}
\usepackage{dblfloatfix}
\useunder{\uline}{\ul}{}
\AtBeginDocument{%
  }

\copyrightyear{2024}
\begin{document}
\title{Context-Aware Trajectory Anomaly Detection}

\author{Haoji Hu, Jina Kim, Jinwei Zhou, Sofia Kirsanova, JangHyeon Lee, and Yao-Yi Chiang}
\affiliation{
  \institution{University of Minnesota, Twin Cities}
  \city{Minneapolis}
  \state{Minnesota}
  \country{USA}}
\email{{huxxx899, kim01479, zhou1909, kirsa002, lee04588, yaoyi}@umn.edu}
\renewcommand{\shortauthors}{Hu et al.}
\begin{abstract}
Trajectory anomaly detection is crucial for effective decision-making in urban and human mobility management. Existing methods of trajectory anomaly detection generally focus on training a trajectory generative model and evaluating the likelihood of reconstructing a given trajectory. However, previous work often lacks important contextual information on the trajectory, such as the agent's information (e.g., agent ID) or geographic information (e.g., Points of Interest (POI)), which could provide additional information on accurately capturing anomalous behaviors. To fill this gap, we propose a context-aware anomaly detection approach that models contextual information related to trajectories. The proposed method is based on a trajectory reconstruction framework guided by contextual factors such as agent ID and contextual POI embedding. The injection of contextual information aims to improve the performance of anomaly detection. We conducted experiments in two cities and demonstrated that the proposed approach significantly outperformed existing methods by effectively modeling contextual information. Overall, this paper paves a new direction for advancing trajectory anomaly detection.

\end{abstract}

\keywords{trajectory anomaly detection, variational autoencoder, contextual information}


\maketitle
\section{Introduction}
Advancements in Global Positioning System (GPS) and sensor technologies have accelerated the generation of human trajectories to an unprecedented pace. Human trajectory data can reflect agents' mobility behavior, providing insights into business strategies or governments. For example, analyzing an agent's routine commute to work and visits to a bar can enable businesses to recommend nearby bars or relevant places. Furthermore, a sudden change in the agent's daily pattern can reflect a habit switch and provide new insights into the individual's behavior or preferences. Thus, trajectory anomaly detection is a pivotal problem in monitoring and analyzing abnormal patterns. Detecting anomalous trajectories can reveal changes in daily routines and provide valuable information for policymakers.


Trajectory anomaly detection has been gaining attention recently~\cite{liu2020online,zhang2023online,chen2024deep}. Existing trajectory anomaly detection borrows ideas from anomaly detection techniques developed for other types of data, such as time series data~\cite{blazquez2021review} and images~\cite{liu2024deep}. These methods transform the trajectory data into a `sequence format' and apply generative models to reconstruct the sequence for anomaly detection. For example,~\cite{liu2020online} introduces a trajectory anomaly detection method that utilizes the detection-via-generation framework with Variational Autoencoder (VAE)~\cite{kingma2013auto}, which is widely employed for detecting anomalies in time-series data~\cite{an2015variational} and images~\cite{yao2019unsupervised}. The underlying idea is that the learned VAE should effectively reconstruct normal data while reconstructing anomalous data is difficult. However, existing studies lack information on the context of trajectory data, which can provide valuable insights for detecting anomalous patterns. Thus, we propose a context-aware trajectory anomaly detection method that incorporates the contextual information of the trajectory data, built upon a representative reconstruction-based anomaly detection method~\cite{liu2020online}. Our goal is to inject context features into trajectories to enable precise reconstruction and generation of trajectories and ultimately improve the performance of anomaly detection.

To enrich the trajectory data, we aim to incorporate (1) auxiliary information (e.g., the identity of an agent, so-called agent ID) and (2) external geographic data, such as Points of Interest (POI) databases. First, regarding auxiliary information, the trajectories with the same points can be normal for one agent, while anomalous for another agent if the trajectory presents an unusual pattern in the person's behavior. For example, a rare visit to New York City for an individual who rarely travels may be flagged as an anomaly, while it could be a regular occurrence for a resident of New York City. The agent ID can help distinguish these two agents. Second, contextual information (e.g., neighboring POIs) can provide important clues to detect anomaly trajectories. For example, an agent's visit to a coffee shop nearby a campus can be a different mobility type from visiting a coffee shop at the airport. Therefore, we leverage the two types of context: the identity of the agent (i.e., agent ID) and the neighboring POIs of trajectories. The proposed method can capture potential connections between trajectories and the context by reconstructing the trajectory conditionally on contextual information. We conducted experiments on the trajectory data in two large cities to demonstrate the effectiveness of the proposed method compared to the existing trajectory anomaly detection method, which did not incorporate any contextual information. 


\section{Problem Statement}
Given a set of agents $\{a_1, a_2, ... a_n\}$, we denote a trajectory of an agent $a_n$ over a long period as a sequence of chronologically 
ordered $m$ points $P^{a_n} = \{p_1^{a_n} \rightarrow p_2^{a_n} \rightarrow ... \rightarrow p_m^{a_n}\}$, where each point 
$p_m^{a_n}$ is represented by GPS coordinates. A subtrajectory of a trajectory is defined as a subsequence of a trajectory with any number of consecutive points. A subtrajectory could be an anomaly if it deviates significantly from the regular pattern in the entire trajectory. Under these definitions, given a set of trajectories from a large group of agents (e.g., from a city), this paper aims to detect the agent whose trajectories contain anomalous subtrajectories. Existing work on trajectory anomaly detection~\cite{liu2020online,zhang2023online} focuses on detecting anomalies for a short trajectory while the trajectory of an agent is much longer than the maximum input length of the model. To mitigate this issue, we propose a method to preprocess the long trajectories into short ones and aggregate the results from short trajectories to obtain the anomaly detection results at an agent level.

\section{Method}
\begin{figure*}[t]
    \centering
    \includegraphics[width=\linewidth]{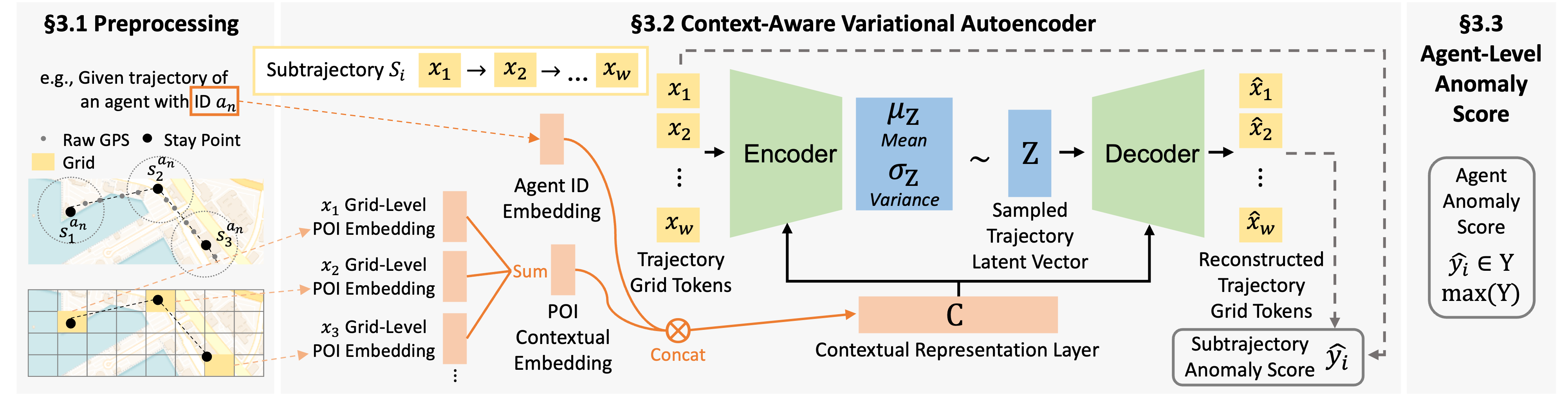}
    \caption{An overview of the proposed method. Section~\ref{sec:preprocess} shows preprocessing steps to process Global Positioning System (GPS) trajectory into stay points and stay points into grid token sequence. 
    Section~\ref{sec:vae} presents how we encode trajectory grid tokens into context-aware Variational Autoencoder (VAE) conditioning on contextual representations, which incorporates agent ID embedding and Points of Interest (POI) contextual embedding. Lastly, Section~\ref{sec:infer} shows the retrieval of an agent-level anomaly score from the subtrajectory anomaly scores.}
    \label{fig:overview}
\end{figure*}

Figure~\ref{fig:overview} shows an overview of the proposed method. Section~\ref{sec:preprocess} presents the procedure to generate the sequence of stay points from a raw GPS trajectory. Next, Section~\ref{sec:vae} shows how we encode preprocessed trajectory grid tokens into context-aware representations and leverage contextual representations to condition VAE. Lastly, Section~\ref{sec:infer} indicates the proposed way of getting agent-level anomaly scores from the anomaly scores of a set of subtrajectories.

\subsection{Trajectory Preprocessing}\label{sec:preprocess}
The original trajectory is represented by the raw GPS points over time. To mitigate the noise in the GPS signal and retain the overall semantics of the trajectory, we apply a commonly used preprocessing method to transform an original trajectory into stay points~\cite{zheng2009mining, hu2022clustering, lin2024unified}. The trajectory record of an agent over a long period can be very long, but subtrajectories could reflect the diverse mobility types of an agent. Thus, we propose strategies to partition the trajectory of an agent $a_n$ into subtrajectories $\mathbf{S}_i$ consists of a sequence of stay points, $\{s_{1}^{a_n}, s_{2}^{a_n}, ..., s_{w}^{a_n}\}$ and expect that different subtrajectories could represent diverse mobility types. Finally, we partition the area of a city into grids and map the stay points of all subtrajectories from all agents into sequences of grid tokens. Thus, we encode the grid token sequence of the $i$-th subtrajectory, $\mathbf{x}_i = \{x_{1}, x_{2}, ..., x_{w}\}$, where $w$ is the sequence length, into the context-aware VAE. Each token corresponds to the grid where a stay point has been mapped.

The Stay Point Detection algorithm~\cite{yue2019detect} extracts stay points from the raw trajectory by partitioning the trajectory with a given time duration threshold and a radius representing the surrounding area. We propose two strategies for partitioning the long trajectory. First, if an agent stays at a location for a long time, we can split the trajectory at this location. The reason is that the agent stays at a location for a long time, indicating that the agent is doing something there. For example, an agent spends a night sleeping at home for several hours, or an agent spends multiple hours working in the company. Second, if an agent spends too long (e.g., 5 hours) on a transition (i.e., from one location to another), we also cut the trajectory at this transition. The reason is that an urban trip should be finished quickly due to the size of the city. A transition that is too long may result from a missing location in the middle. In this way, the long trajectory of an agent over a long period is partitioned into short subtrajectories with diverse mobility types. After partitioning long trajectories into a set of subtrajectories, we map stay-point sequences into grid token sequences.





\subsection{Context-Aware Variational Autoencoder}
\label{sec:vae}
We develop the proposed method by employing contextual information (e.g., the agent ID of the trajectory and the surrounding POIs) to guide reconstruction for anomaly detection upon existing VAE-based method~\cite{liu2020online}. After the preprocessing steps, the stay points sequence is transferred into grid token sequences for VAE to reconstruct (Figure~\ref{fig:overview}). In addition, the agent ID and POI contextual information become the conditions to guide the reconstruction. We generate the anomaly score by comparing the reconstructed and original token sequences.

VAE-based methods such as~\cite{liu2020online} can reconstruct any given token sequence and compare the reconstructed token sequence with the input token sequence to obtain the anomaly score. Although the Gaussian Mixture Model (GMM) could be used as the prior latent variables and allow the VAE model~\cite{liu2020online} to model the multinomial distribution, we argue that the diversity of agents' trajectory patterns is too complex to capture with the GMM. Thus, we propose using personalized, contextual information to capture these personalized details and guide the generative model. Particularly, we use a conditional VAE instead of a vanilla VAE to obtain the anomaly score of a trajectory. Our proposed model consists of two main steps: 1) contextual representation generation and 2) context-aware token sequence reconstruction. 

\paragraph{\textbf{Contextual Representation Generation}} We develop a unified contextual layer for encoding contextual information as below. 

\begin{itemize}
    \item{\textbf{Agent Embedding:}} Several agents typically have various moving behaviors. To model the personalized pattern at the agent level, we propose to use an agent embedding that could map an agent ID into a vector as an agent embedding. A similar idea has been widely used in recommendation systems to model the personalized pattern for people~\cite{he2017neural}. Agents with similar behaviors are expected to produce similar agent embedding, which the model learns in an end-to-end fashion.
    
    \item {\textbf{POI Contextual Embedding:}} 
    Surrounding POIs of trajectories can provide insights into mobility types of trajectories. POIs consist of textual data (e.g., name and categories) and spatial data (e.g., geocoordinates). The goal here is to generate contextual POI embeddings that capture the characteristics of surrounding POIs, ensuring that POIs within the same category are represented by distinct embeddings. To achieve this, we extend SpaBERT~\cite{li2022spabert} to leverage pretrained language models to contextualize the semantic meanings and spatial and topological relationships of POIs in relation to their neighboring POIs and polygons of interest. After generating the contextual POI embeddings, we cluster all learned embeddings and assign a cluster type to each POI as a new contextualized category, replacing the original POI categories. A grid-level POI embedding is then represented by a count vector indicating the contextual category of POIs within each grid. Finally, we sum all grid-level POI embeddings from the grid sequences to obtain a final contextual POI embedding that represents a subtrajectory. The final contextual POI embedding is used to guide the reconstruction of the token sequence.
\end{itemize}

\paragraph{\textbf{Context-Aware Token Sequence Reconstruction}}
Figure~\ref{fig:overview} presents the architecture of the proposed context-aware generative model. Given a subtrajectory's sequence of grid tokens, we leverage the contextual embedding-guided conditional generative model to reconstruct the input grid token sequence. The main idea of our context-aware VAE anomaly detection is that, for a given grid token sequence input of $i$-th subtrajectory $\mathbf{x}_i = \{x_{1}, x_{2}, ..., x_{w}\}$ of length $w$ and the corresponding contextual representation $\mathbf{c}_i$, the corresponding latent variable $\mathbf{z_i}$ is  

\begin{equation}
    \label{eq:1}
    \mathbf{\mu}_{\mathbf{z}_i}, \mathbf{\sigma}_{\mathbf{z}_i} = f_{\theta}(\mathbf{z}_i|\mathbf{x}_i, \mathbf{c}_i)
\end{equation}

\begin{equation}
    \label{eq:2}
    \mathbf{z_i} \sim \mathcal{N}(\mathbf{\mu}_{\mathbf{z}_i}, \mathbf{\sigma}_{\mathbf{z}_i})
\end{equation}

where $f_{\theta}(\mathbf{z}_i|\mathbf{x}_i, \mathbf{c}_i)$ is the encoder neural network and models the probability $q_{\phi}(\mathbf{z}_i|\mathbf{x}_i, \mathbf{c}_i)$. 

Then, the reconstruction probability is calculated using the Monte Carlo estimate with $L$ sampling pairs $(\mathbf{\mu}_{\mathbf{\hat{x}}_{i,l}}, \mathbf{\sigma}_{\mathbf{\hat{x}}_{i,l}})$: 

\begin{align}
\small
\label{eq:3}
\mathbf{\mu}_{\mathbf{\hat{x}}_{i,l}}, \mathbf{\sigma}_{\mathbf{\hat{x}}_{i,l}} &= g_{\phi}(\mathbf{x}_i|\mathbf{z}_{i,l}, \mathbf{c}_i) \\
E_{q_{\phi}(\mathbf{z}_i|\mathbf{x}_i)}[\log p_{\theta}(\mathbf{x}_i|\mathbf{z}_i)] &= \frac{1}{L}\sum^{L}_{l=1}p_{\theta}(\mathbf{x}_i|\mathbf{\mu}_{\mathbf{\hat{x}}_{i,l}}, \mathbf{\sigma}_{\mathbf{\hat{x}}_{i,l}})
\end{align}

where the $g_{\phi}(\mathbf{x}_i|\mathbf{z}_{i,l}, \mathbf{c}_i)$ is the decoder neural network to model probability $p_{\theta}(\mathbf{x}_i|\mathbf{z}_i, \mathbf{c}_i)$. $E_{q_{\phi}(\mathbf{z}_i|\mathbf{x}_i)}[\log p_{\theta}(\mathbf{x}_i|\mathbf{z}_i)]$ is the likelihood to reconstruct the input sequence and $\mathbf{\hat{y}_i}=1-E_{q_{\phi}(\mathbf{z}_i|\mathbf{x}_i)}[\log p_{\theta}(\mathbf{x}_i|\mathbf{z}_i)]$, interpreted as an anomaly score. The lower the probability of reconstruction, the more likely the time step $i$ is an anomaly. 

The objective function to train the context-aware VAE network is the evidence lower bound (ELBO) over all the sequences as follows:
\begin{equation}
\small
\label{eq:4}
\mathcal{L} \text{=} \sum_i E_{q_{\phi}(\mathbf{z_i}|\mathbf{x}_i, \mathbf{c}_i)}\log p_{\theta}(\mathbf{x}_i|\mathbf{z_i}, \mathbf{c}_i) \text{-}D_{KL}[q_{\phi}(\mathbf{z_i}|\mathbf{x}_i, \mathbf{c}_i)||p(\mathbf{z_i})]
\end{equation}

\subsection{Agent-Level Anomaly Inference}
\label{sec:infer}
Our context-aware generative model only provides the anomaly score for each grid-token sequence. To obtain the anomaly score for the agent level, we propose to infer the agent anomaly score based on the token sequence anomaly scores. Even though multiple token sequences could be detected as anomalies, we argue that the most suspicious part of the trajectory could reflect the suspicious level of the agent. Thus, we use the maximum anomaly score over all the anomaly scores from all grid-token sequences of an agent (i.e., all subtrajectories of an agent) as the agent-level anomaly score.


\section{Experiments}
\paragraph{\textbf{Experiment Settings.}}
We evaluate our methods using simulated trajectory data from two cities. We partition the whole data into two halves along the temporal dimension for training and testing, respectively. The grid size is set to 500 meters $\times$ 500 meters (chosen under hyperparameter search). Table~\ref{tab:stat} shows the descriptive statistics of trajectory data after preprocessing. 

\begin{table}[h]
\centering
\caption{Descriptive statistics of datasets.}
\label{tab:stat}
    \scalebox{0.8}{%
    \begin{tabular}{|l|ccccc|}
    \hline
    \textbf{Location}          & \textbf{\#Agents} & \textbf{\#POIs} & \makecell{\textbf{\# Training}
    \\ \textbf{Grid-Token} \\ \textbf{ Sequences}} & \makecell{\textbf{\# Test} \\ \textbf{Grid-Token}  \\ \textbf{Sequences}} & \makecell{\textbf{\# Grid-Token} \\ \textbf{Sequence} \\ \textbf{Length} } \\ \hline
    City 1       & 152,586  &595,125       & 2,745,662                   & 2,741,000  & [2, 18]                                 \\ \hline
    City 2     & 499,875    &80,956     & 8,807,570                   & 1,497,371   & [2, 32]                                 \\ \hline
    \end{tabular}%
    }
\end{table}

Anomaly detection is a binary label problem, and we can employ the F1 score as an evaluation metric. However, the F1 score depends on the anomaly threshold. Area Under the Receiver Operating Characteristic Curve (AUC-ROC) is another commonly employed metric for binary label problems. However, the data distribution in anomaly detection is highly imbalanced, as positive instances (i.e., anomalies) are rare compared to negative instances (i.e., normal). Thus, AUC-ROC can be dominated by the number of negative instances and produce optimistic results. To address these issues, we employ the precision-recall curve as our evaluation metric, which is recommended for highly skewed domains where ROC curves may provide an overly optimistic view of performance~\cite{paula2016survey}.

\paragraph{\textbf{Experiment Results.}}
Table~\ref{tab:results} shows the experimental results. Using both POI contextual embedding and agent ID embedding shows the best performance for two locations. The results imply several meanings as follows. First, by comparing the method with no context to using POI categories directly as context, POI categories cannot guarantee performance improvement. The reason is that the frequencies of POI categories may not fully capture the spatial characteristics of grids, meaning that grids having similar distribution of POI categories can still have diverse characteristics. For example, starting a trajectory from school and stopping in a school encodes the frequencies of the same POI category. This may not provide enough personalized information for an agent, resulting in low performance. Second, by comparing the method with no context to employing our new contextualized category of POIs (i.e., POI contextual embedding setting), involving our contextualized category of POIs can perform better than using the original POI categories. We believe that, as the POI contextual embeddings capture the semantic meanings and spatial relationships among neighboring POIs, incorporating the learned POI contextual embedding becomes discriminative even for POIs of the same categories. Regarding `City 2', the performance of using no context is higher than the one using POI categories or POI contextual embedding. We believe that the low performance is due to the lack of POIs for the areas that trajectories pass by (i.e., the inconsistent gaps between \#Agents and \#POIs in Table~\ref{tab:stat} between two cities). Third, by comparing no context with the agent ID embedding setting, agent ID could consistently improve the performance. This suggests the importance of agent ID in reflecting personalized information at the agent level. Fourth, by comparing no context with the combination of POI contextual embedding and agent ID embedding, the combined context could achieve the best performance. This indicates that personalized information from the new contextualized categories of POIs and agent IDs can complement each other and provide a comprehensive context for personalization. This combination can further improve the performance. 

\begin{table}[h]
\small
\centering
\caption{Experiment results on anomaly detection.}
\label{tab:results}
\resizebox{0.48\textwidth}{!}{%
\begin{tabular}{|l|c|c|c|c|c|}
\hline
\textbf{Location}    & \makecell{\textbf{No} \\ \textbf{context}} & \makecell{\textbf{POI} \\ \textbf{Categories}} & \makecell{\textbf{POI} \\ \textbf{Contextual} \\ \textbf{Embedding}}         & \textbf{Agent ID}&  \makecell{\textbf{POI} \\ \textbf{Contextual} \\ \textbf{Embedding} + \\  \textbf{Agent ID}} \\ \hline
City 1       & 0.0457     & 0.0531         & 0.0835          & \underline{0.1609}    & \textbf{0.2212}            \\ \hline
City 2     & 0.3675     & 0.3064         & 0.3122          & \underline{0.4618}    & \textbf{0.4725}            \\ \hline
\end{tabular}%
}
\end{table}

\begin{figure}[h]
    \centering
    \includegraphics[width=0.95\linewidth]{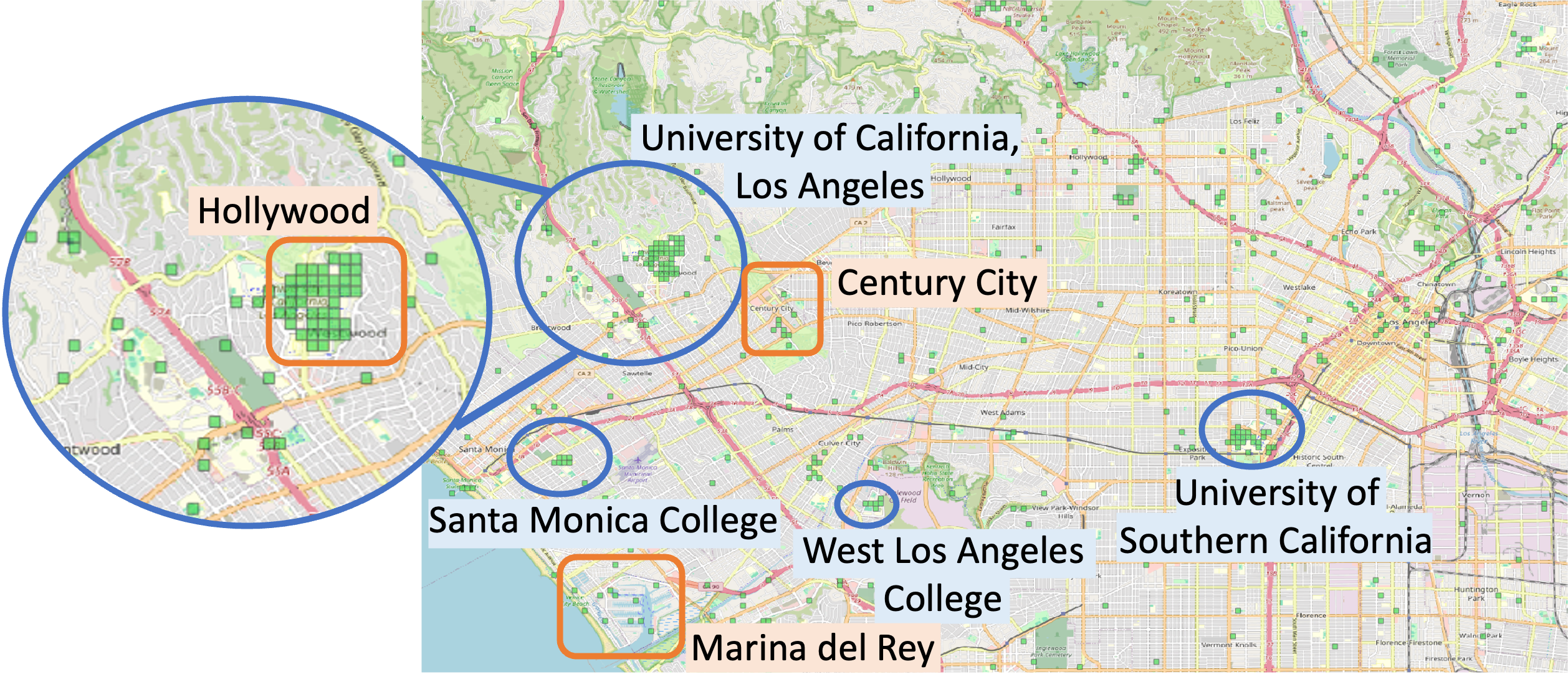}
    \caption{A visualization of grids from a specific cluster label generated by clustering the learned contextual embeddings of Points of Interest (POI).}
    \label{fig:embedding}
    \Description{}
\end{figure}

\paragraph{\textbf{Case Study of POI Contextual Embeddings.}}
Our learned POI contextual embeddings represent the frequency of clusters within each grid, and we can interpret the cluster labels as a new latent semantic type. To demonstrate the effectiveness, we learn the POI contextual embedding in Los Angeles and visualize a particular latent semantic type in Figure~\ref{fig:embedding} by highlighting the grids with the same color (i.e., green-colored grids). The grids with the same color appear not only over the campus areas (i.e., blue-colored text labels) but also in Century City, near Hollywood, and in the bottom-left corner at Marina del Rey (i.e., orange-colored text labels). Each location is a vibrant area in Los Angeles that attracts residents, tourists, and businesses, often hosting cultural activities and events. The qualitative analyses of the learned latent semantic type indicate that the learned semantic type can capture the distinct characteristics of areas, which provides a more comprehensive context for anomaly detection than the original POI categories.


\section*{Acknowledgments}
Supported by the Intelligence Advanced Research Projects Activity (IARPA) via the Department of Interior/Interior Business Center (DOI/IBC) contract number 140D0423C0033. The U.S. Government is authorized to reproduce and distribute reprints for Governmental purposes notwithstanding any copyright annotation thereon. Disclaimer: The views and conclusions contained herein are those of the authors and should not be interpreted as necessarily representing the official policies or endorsements, either expressed or implied, of IARPA, DOI/IBC, or the U.S. Government.

\end{document}